\documentclass[letterpaper]{article} % DO NOT CHANGE THIS
\usepackage{aaai25}  % DO NOT CHANGE THIS
\usepackage{times}  % DO NOT CHANGE THIS
\usepackage{helvet}  % DO NOT CHANGE THIS
\usepackage{courier}  % DO NOT CHANGE THIS
\usepackage[hyphens]{url}  % DO NOT CHANGE THIS
\usepackage{graphicx} % DO NOT CHANGE THIS
\def\UrlFont{\rm}  % DO NOT CHANGE THIS
\usepackage{natbib}  % DO NOT CHANGE THIS AND DO NOT ADD ANY OPTIONS TO IT
\usepackage{caption} % DO NOT CHANGE THIS AND DO NOT ADD ANY OPTIONS TO IT
\usepackage{algorithm}
\usepackage{algorithmic}
\usepackage[utf8]{inputenc} % allow utf-8 input
\usepackage[T1]{fontenc}    % use 8-bit T1 fonts
\usepackage[mathscr]{eucal}
\usepackage{comment}
\usepackage{url}            % simple URL typesetting
\usepackage{booktabs}       % professional-quality tables
\usepackage{amsfonts}       % blackboard math symbols
\usepackage{nicefrac}       % compact symbols for 1/2, etc.
\usepackage{microtype}      % microtypography
\usepackage{xcolor}         % colors
\usepackage{amsmath}
\usepackage{amssymb}
\usepackage{array}
\usepackage{mathtools}
\usepackage{amsthm}
\usepackage{subcaption}
\usepackage{graphicx}
\usepackage{caption}
\usepackage{enumitem}   
\usepackage{flexisym}
\let\oldcdot\cdot
\usepackage{breqn}
\let\cdot\oldcdot
\newtheorem{theorem}{Theorem}[section]
\newtheorem{lemma}[theorem]{Lemma}
\newtheorem{asu}{Assumption}

\DeclarePairedDelimiter\floor{\lfloor}{\rfloor}
\makeatletter
\newcommand{\anabla}{\mathord{\mathpalette\raise@half\bigtriangledown}}
\newcommand\raise@half[2]{%
	\raisebox{\depth}{$\m@th#1#2$}%
}
\usepackage{newfloat}
\usepackage{listings}
\DeclareCaptionStyle{ruled}{labelfont=normalfont,labelsep=colon,strut=off} % DO NOT CHANGE THIS
\floatstyle{ruled}
\newfloat{listing}{tb}{lst}{}
\floatname{listing}{Listing}
\title{Non-Convex Optimization in Federated Learning via Variance Reduction \\
and Adaptive Learning}
\author{
    Dipanwita Thakur\textsuperscript{\rm 1},
    Antonella Guzzo\textsuperscript{\rm 1},
    Giancarlo Fortino\textsuperscript{\rm 1}
    Sajal K. Das\textsuperscript{\rm 2}
}
\affiliations{
     \textsuperscript{\rm 1}University of Calabria, Italy\\
    \textsuperscript{\rm 2}Missouri University of Science \& Technology,  USA\\
    dipanwita.thakur@unical.it, antonella.guzzo@unical.it, giancarlo.fortino@unical.it, sdas@mst.edu
}

\usepackage{bibentry}
\begin{document}

\maketitle

\begin{abstract}

This paper proposes a novel federated algorithm that leverages momentum-based variance reduction with adaptive learning to address non-convex settings across heterogeneous data. We intend to minimize communication and computation overhead, thereby fostering a sustainable federated learning system. We aim to overcome challenges related to gradient variance, which hinders the model's efficiency, and the slow convergence resulting from learning rate adjustments with heterogeneous data. The experimental results on the image classification tasks with heterogeneous data reveal the effectiveness of our suggested algorithms in non-convex settings with an improved communication complexity of $\mathcal{O}(\epsilon^{-1})$ to converge to an $\epsilon$-stationary point - compared to the existing communication complexity $\mathcal{O}(\epsilon^{-2})$ of most prior works.  The proposed federated version maintains the trade-off between the convergence rate, number of communication rounds, and test accuracy while mitigating the client drift in heterogeneous settings. The experimental results demonstrate the efficiency of our algorithms in image classification tasks (MNIST, CIFAR-10) with heterogeneous data.
\end{abstract}

\section{Introduction}
Federated learning (FL) is a distributed machine learning technique allowing multiple devices to train a model collaboratively without sharing their data with a central server\cite{McMahan2017}. Traditional federated learning architecture consists of a global server and several local devices. The global server, also known as the federated server, generates a global model. The global model is sent to the local clients to process the locally generated data. The parameters used in the local models are sent to the global server, and the global server aggregates the parameters and sends back the updated global model to the local devices to enhance the performance of the local models using locally generated data. Usually, devices are heterogeneous. However, there are several issues related to the traditional federated learning. 

First, several global rounds are performed in a traditional FL framework to converge with non-independent, Identically Distributed (non-IID) client datasets and high communication costs per round. Hence, it enhances the communication complexity.  
Second, the number of clients participating in each iteration also affects the performance of the traditional FL. Every client can't participate in each round due to slow convergence for tuning the learning rate. Tuning the learning rate also enhances the computational complexity. A lower number of client participants increases the number of global rounds due to the non-IID data as each local device processes its personalized data. Third, the training data are dispersed widely among many devices, and there is a complicated connection time between each device and the central server. The slow communication is a direct result, which inspired FL algorithms with high communication efficiency.
Fourth, non-convex optimization is among the most vital techniques in contemporary machine learning (e.g., neural network training). Convergence in non-convex settings is still challenging in FL settings with non-IID data.  
\par
Cross-silo and cross-device are two different settings of FL. The cross-silo environment is associated with a limited number of dependable clients, usually businesses like banks or medical facilities. On the other hand, there could be a very big number of clients in the cross-device federated learning environment—all 6.84  billion active smartphones, for instance. Because of this, we might never, ever go through all of the clients' data during training in that particular scenario. Additional characteristics of the cross-device environment include resource-poor clients interacting via a very unstable network. The combination of this setting's fundamental elements creates special difficulties that do not exist in the cross-silo setting. In this paper, we consider the more difficult cross-device scenario. Notably, recent developments in FL optimization, including FedDyn \cite{Acar2021FedDyn}, SCAFFOLD \cite{Karimireddy2020b}, and \cite{Wu2023} are no longer relevant because they were created for cross-silo environments.
\par
The computation related to federated learning design, despite the model used for training, is directly related to energy consumption. Hence, we need an optimized FL algorithm that gives convergence guarantees. Due to the issues mentioned above, designing an optimized FL that converges is challenging. Several FL algorithms proposed so far mainly focus on non-iid data and communication efficiency. To develop the convergence guarantee for the FL algorithm, the researchers assumed either iid data or the participation of all the devices\footnote{Throughout the article, we refer to network things like nodes, clients, sensors, or organizations as ``devices.''}.
However, these assumptions are not feasible in real scenarios. In the real world, it is necessary to handle heterogeneous (non-IID) data to mitigate the ``client-drift'' problem. Several optimization algorithms are proposed in the FL literature to mitigate the client drift issue. However, very few are considered in both the non-convex and non-IID settings and provide the convergence analysis for the same. 

\par
This paper aims to decrease the number of communication rounds, which ultimately reduces energy consumption. According to \cite{McMahan2017}, FL settings are more prone to communication overhead than computation overhead. The objective is to increase computation cost to minimize the communication rounds to train a model. This can be achievable either by increasing parallelism or increasing the computation of each device. Participation of more clients with higher computation can reduce the number of communication rounds \cite{mcmahan2023communicationefficient}. Most FL algorithms consider a fixed learning rate for all clients in each iteration. However, not all parameters will benefit from this uniform learning rate, which could lead to slow convergence due to high computation for tuning of learning rate. While certain parameters may require smaller adjustments to avoid overshooting the optimum value, others may require more frequent updates to speed up convergence.

To tackle this, algorithms with adjustable learning rates were created. With these methods, the algorithm can traverse the optimization landscape more quickly by modifying the learning rate of each parameter according to its past gradients. 
Even though several FL techniques have been put forth, their undesirable convergence behavior prevents them from significantly lowering communication costs by avoiding frequent transmission between the central server and local worker nodes. Numerous factors contribute to its cause, including (1) client drift \cite{Karimireddy2020b}, in which local client models approach local rather than global optima; (2) lack of adaptivity as stochastic gradient descent (SGD) based update \cite{Reddi2021} and (3) large training iterations to converge as a model parameter and training dataset sizes keep growing. Even with these new developments, most research is devoted to fixing client drifts \cite{Karimireddy2020b, Khanduri2021STEMAS, Xu2022}. Not all problems can be resolved by the FL system that is in place now.

\par
Several optimization techniques are available for gradient descent to reduce the loss function and converge the algorithm. Momentum-based updates and adaptive learning rates are the two popularly used optimization techniques for gradient descent. The deep learning algorithms normally used as a model in the FL frameworks are of nonconvex settings. Momentum-based updates are used to overcome local minima and accelerate convergence, and the adaptive learning rates are used to improve the convergence speed and accuracy of the algorithm. Traditional optimizers, such as Adaguard, RMProp, and Adam, are used in adaptive federated learning. However, the structure of adaptive FL systems is challenging because the local device moves in diverse directions, and the global servers cannot be updated often in the FL scenario. Convergence problems may arise from the adaptive FL method's flawed architecture \cite{Chen2020TowardCE}. FedAdagrad, FedYogi, and FedAdam are the federated versions of adaptive optimizers first proposed in \cite{Reddi2021}. However, this analysis is only valid when $\beta1 = 0$. Therefore, it is unable to make use of momentum. 
To overcome this issue, MimeAdam is proposed in \cite{Karimireddy2020a}, which applies server statistics locally. However, MimeAdam needs to calculate the entire local gradient, which may not be allowed in real life. 

FedAMS is a more recent proposal \cite{wang2023communicationefficient} that offers complete proof but does not increase the pace of convergence. FedAdagrad, FedYogi, FedAdam, and FedAMS have sample convergence rates of $\mathcal{O}(\epsilon^{-4})$ overall, which is not better than FedAvg. They also need an additional global learning rate to tune simultaneously. In \cite{Rostami2023}, the authors used a stochastic and variance-reduced technique called Stochastic Variance Reduced Gradient (SVRG) for local updates. This method calculated the full batch gradient of each participating device at each iteration which ultimately enhanced the computational complexity. In \cite{Wu2023}, the authors proposed an effective adaptive FL algorithm (FAFED) in cross-silo settings using the momentum-based variance-reducing technique. FAFED shows the fast convergence with the best-known sample complexity $\mathcal{O}(\epsilon^{-3})$ and communication complexity $\mathcal{O}(\epsilon^{-2})$. However, the non-convex setting required for deep neural network (DNN) algorithms is not addressed. Moreover, DNN requires higher GPU memory while analyzing the images. Hence, large batch sizes or calculation of gradient checkpoints increase the computational complexity with slow convergence. 
\par

Based on the issues mentioned above, we propose a new FL algorithm in this paper using the adaptive learning rate with variance reduction of momentum SGD in non-convex settings.
Specifically, we suggest a general energy-efficient, optimized FL framework using the concept of reduced communication rounds in which (1) the devices train over several epochs using an adaptive learning rate optimizer to minimize loss on their local data and (2) the server updates its global model by applying a momentum-based gradient optimizer to the average of the clients' model updates, while the clients train over several epochs using a client optimizer to minimize loss on their local data. The proposed method combines momentum-based variance reduction updates and adaptive learning rate optimization techniques without increasing client storage or communication costs for local and global updates.

\par
\textbf{Contributions.} our major contributions are as follows.
\begin{itemize}
    
    \item We design a federated learning algorithm where momentum-based variance reduction is used for global updates in addition to the momentum-based local updates with adaptive learning. The design of the proposed FL algorithm is aimed at addressing two main issues in order to speed up the convergence of Federated Learning (FL) and another issue to mitigate the high computation with non-IID setting for non-convex problems. These issues are the high variances associated with: (i) the noise from local client-level stochastic gradients, and (ii) the global server aggregation step due to varying client characteristics when multiple local updates are present. A high computation issue is associated with tuning the learning rate. We can tackle these issues separately by reducing the variance in the global server update and the local client updates using both global and local momentum with adaptive learning.
    
    \item We give convergence analysis in generic non-convex settings and design new federated optimization methods compatible with multiple devices using the proposed FL algorithm. The convergence analysis shows that the proposed FL algorithm converges to an $\epsilon-$ stationary point for smooth non-convex problems with a non-IID setting with improved communication complexity of $\mathcal{O}(\epsilon^{-1})$ compared with state-of-the-art approaches.
    
    \item Experimental results with non-IID data in non-convex settings reveal the effectiveness of our suggested algorithm which encourages communication efficiency and subtly reduces client drift under heterogeneous data distribution.

\end{itemize}

%%%%%%%%%%%%%%%%%%%%%%%%%%%%%%%%%%%%%%%%%%%%%%%%
\section{Related Works}
While several FL algorithms have been proposed, most of them are related to communication efficiency and data privacy. Our work focuses on communication-efficient and early convergent FL algorithms. 
Federated Averaging (FedAvg) \cite{McMahan2017} is the first and most popularly used FL algorithm. However, FedAvg fails to guarantee theoretically with non-IID in the convex optimization setting. FedAvg does not completely address the fundamental issues related to heterogeneity, despite the fact that it has empirically proven successful in diverse environments. FedAvg in the setting of systems heterogeneity typically drops devices that are unable to compute E epochs within a given time window, rather than allowing participating devices to undertake varying amounts of local work based on their underlying systems restrictions. To address this issue, several regularization approaches \cite{Acar2021FedDyn, Gao2022, Karargyris2023, Mendieta2022} are used to enforce local optimization. FedProx \cite{li2020federated} is another FL algorithm without assuming IID data and the participation of all the client devices. 

Specifically, FedProx exhibits noticeably more accurate and steady convergence behavior than FedAvg in extremely heterogeneous situations. The FedProx methodology relies on a gradient similarity assumption. The FedProx adds a proximal term for each local objective to prove the convergence. However, several presumptions are essential to implement FedProx in realistic federated contexts. The SCAFFOLD approach \cite{Karimireddy2020b} achieves convergence rates independent of the degree of heterogeneity by utilizing control variates to lessen client drift. Although the approach works well for cross-silo FL, it cannot be used for cross-device FL because it requires clients to keep their states constant over rounds. A similar approach, FedDyn \cite{Acar2021FedDyn}, was proposed to reduce communication rounds using linear and quadratic penalty terms. Despite their best efforts, FedAvg's performance is not entirely understood. However, it is proved from the above that client drift is one of the major issues for performance degradation in FL. It introduced control variate to adjust local updates in order to address the client drift issue. Furthermore, MOON \cite{Li2021MOON} suggested using model contrastive learning to use the similarity between local and global model representations to rectify the local training. Various research attempts to enhance FedAvg in various ways. For example, in  \cite{Wang2021FedNova}, the authors explained that the objective inconsistency is the reason behind the slow convergence and also provided the converge analysis of the proposed and previous methods. On the other hand, FedMa \cite{Wang2020FedMa} was proposed to enhance the convergence rate by matching and averaging hidden elements and also reduce the communication cost.
\par
Compared with existing works on FL \cite{McMahan2017}, \cite{li2020federated}, \cite{Karimireddy2020b}, this paper uses two different optimization methods to reduce the communication rounds to make the proposed FL algorithm an energy-efficient one. In contrast to state-of-the-art FL algorithms, the client devices optimize their local models using an adaptive learning rate optimizer to minimize loss on their local data, and the server updates its global model by applying a momentum-based gradient optimizer to the average of the clients' model updates.  However, none of them studies the variance reduction with adaptive learning in the federated environment. Table \ref{tab:rw} the comparison of our proposed method with state-of-the-art methods which considered non-IID settings and non-convex functions.

\begin{table*}[ht]
\centering
\begin{tabular}{cccccc}
\hline
     \textbf{Reference}& \textbf{Client Participation} &\textbf{Compression?} & \textbf{BCD?} & \textbf{ALR?} & \textbf{T}\\
     \hline
      \cite{Koloskova2020} & Full & No & Yes & No & $\mathcal{O}(\frac{1}{N\epsilon^{2}})$\\
      \hline
      \cite{Haddadpour2020FederatedLW} & Full & Yes & Yes & No & $\mathcal{O}(\frac{1}{N\epsilon^{2}})$\\
      \hline
      \cite{Khanduri2021STEMAS} & Full & No & Yes & No & $\mathcal{O}(\frac{1}{N\epsilon^{1.5}})$\\      \hline
      \cite{Karimireddy2020a} &   Partial & No & Yes & No & $\mathcal{O}(\frac{1}{\sqrt{R\epsilon^{1.5}}})$\\
      \hline
      \cite{Karimireddy2020b} & Partial & No & Yes & No & $\mathcal{O}(\frac{1}{R\epsilon^{1.5}})$\\
      \hline
      \cite{Das2022faster} & Partial & Yes & No & No & $\mathcal{O}(\max{(\sqrt\frac{\alpha}{N}, \frac{1}{\sqrt{R}})\frac{1}{\epsilon^{1.5}}})^{*1}$\\
      \hline
      Ours & Partial & No & No & Yes & $\mathcal{O}(\frac{1}{\epsilon^{1}})$\\
      \hline
\end{tabular}
\caption{Parameter description in the SOTA and our proposed FL algorithms: Here, N is the total number of clients, P is the number of participating clients in each round, either N=P (Full) or P < N (Partial), compression denoted the either compressed communication or not, BCD denotes the bounded client dissimilarity assumptions, ALR denotes the use of adaptive learning rate or not and T denotes the number of gradient updates to achieve $\mathbb{E}[\lvert f{\omega} \rvert ^{2}] \le \epsilon$ on smooth non-convex settings}
\label{tab:rw}
  \end{table*}
Algorithms \ref{alg:cup} and \ref{alg:sup} depict the federated version for clients and server updates using momentum-based variance reduction with the adaptive learning rate.

%%%%%%%%%%%%%%%%%%%%%%%%%%%%%%%%%%%%%%%%%%%%%%%%%%
\section{Preliminaries}
Focusing on non-convex settings, our objective is to solve an optimization problem with the help of the functions
\begin{equation}
\begin{aligned}
\min_{x \in \mathbb{R}^d} f(\omega) \\ 
\textrm{where} f(\omega) = \frac{1}{N}\sum_{i=1}^{N}{F_{i}(\omega)}
\end{aligned}
\label{eq:eq1}
\end{equation}

$F_i(\omega)=\mathbb{E}_{x \sim \mathcal{D}_i}\mathcal{L}_i(x ; \omega)$, which gives the loss of the prediction on data sample $(x_i, y_i)$ with model parameters $\omega$, $\omega \in \mathcal{Z}$ and $\mathcal{D}_i$ denotes the data distribution for the $i^{\text{th}}$ client. When the distributions $\mathcal{D}_i$ are different among the clients, it is known as the heterogeneous data setting. Each $F_i(\omega)$ is differentiable, possibly non-convex satisfies L-Lipschitz continuous gradient (i.e, L-smooth) for some parameter $L>0$. For each client $i$ and $\omega$, the true gradient $\bigtriangledown F_i(\omega)$ is assumed to have an unbiased stochastic gradient $g_i(\omega)$. The loss on an example is denoted by $f(x, \omega)$, and the model's training loss is denoted by $F$. We can derive sample functions $f(\cdot, x)$ such that $\mathbb E[f(\cdot, x)] = F(\cdot)$, where $x$ denotes a batch of data samples. 
Problem \ref{eq:eq1} encompasses a wide range of machine learning and imaging processing issues, such as neural network training, non-convex loss ERM (empirical risk minimization), and many more.
As per the traditional standard for non-convex optimization, our attention is directed towards algorithms that can effectively identify an approximate stationary point x that satisfies $\lVert \bigtriangledown f(x)\rVert^2 \le \epsilon $.

Furthermore, we also assume the following. \\
\begin{asu} \label{asu1}
\textit{\textbf(Smoothness)}. The function $F_i$ is $L_1$-smooth if its gradient is $L_1$-Lipschitz, that is $\lVert \bigtriangledown F_i(x) - \bigtriangledown F_i(y) \rVert   \le L\lVert x-y \rVert$ for all $x,y \in  \mathbb{R}_d$. We also have: $f(y) \le f(x) + \langle \bigtriangledown f(x), y-x \rangle + \frac{L}{2} \lVert y - x\rVert^2$.
\end{asu}
\begin{asu} \label{asu2}
\textit{\textbf(Bounded Variance)}. The function $F_i$ have $\sigma$-bounded (local) variance i.e., $\mathbb{E}[ \lVert \bigtriangledown[f_i(x,\omega)]_j - [\bigtriangledown F_i(x)]_j \rVert ^ 2] \le \sigma ^2$ for all $x \in \mathbb{R}^d$, $j \in [d]$ and $i \in [n]$. Moreover, we assume that the global variance is also bounded, $(1/n)\sum_{i=1}^{n} \lVert \bigtriangledown [F_i(x)]_j - [f(x)]_j \rVert^2 \le \bigtriangledown^2{g,j}$ for all $x \in \mathbb{R}^d$, $j \in [d]$. 
\end{asu}
\begin{asu}\label{asu3}
\textit{\textbf(Bounded Gradients)}. The function $f_i(x,\omega)$ have $G$-bounded gradients i.e., for any $i \in [n], x \in \mathbb{R}^d$ and $\omega \in \mathcal{Z}$, $\lVert[\bigtriangledown f_i(x, \omega)]_j\rVert \le G$, $\forall j \in [d]$.
\end{asu}
\begin{asu}\label{asu4}
\textit{\textbf(Unbiased Gradients)}. Each component function $f_i(x;\omega)$ computed at each client node is unbiased $\forall x_i \sim \mathcal{D}_i$, $i \in [N]$ and $x \in \mathbb{R}^d$. 
\end{asu}
\begin{comment}
e\begin{asu}\label{asu5}
		\textit{\textbf(Heteogeneity)}. In the stochastic non-convex settings, the heterogeneity in federated learning is measured by 
		$\lVert \bigtriangledown [F_i(x)]_j - \bigtriangledown[f(x)]_j \rVert^2 \le \sigma^2_{G}$, where $\sigma_{G}$ is fixed which represents the upper bound of the dataset heterogeneity. The data heterogeneity can represent the bounded stochastic gradient assumption s.t 
		$\lVert \bigtriangledown [F_i(x)]_j - \bigtriangledown [f(x)]_j \rVert^2 \le 2 \lVert \bigtriangledown [F_i(x)]_j \rVert^2 +  2 \lVert \bigtriangledown [f(x)]_j \rVert^2 \le 4dG^2$
	\end{asu}nt...
\end{comment}

\begin{asu}\label{asu5}
    \textit{\textbf{Non-negativity}}. Considering each $f_i(\omega)$ is non-negative and hence, $f_i^* = min f_i(\omega) \ge 0$. While computation, most of the loss functions are usually positive. In case of negative, we add some constant value to make it positive.
\end{asu}
\textit{\textbf{Definition I}} ($\epsilon$-stationary Point). $\epsilon$-stationary point, x satisfies $\lVert \bigtriangledown f(x) \rVert \le \epsilon$. Furthermore, in $t$ interactions, a stochastic algorithm is considered to reach an $\epsilon$-stationary point if $\mathbb E[\lVert \bigtriangledown f(x) \rVert] \le \epsilon$, where the expectation is over the algorithm's randomness up to time $t$. \\

%\textit{\textbf{Definition II}} (Computational Complexity/ Sample Complexity). In an Incremental First-order Oracle (IFO) framework, given a sample $x_{i}$ at the $i\textsuperscript{th}$ client and iterating $\omega$, the oracle returns $(f_i(x, \omega)]_j), \bigtriangledown f_i(x, \omega)_j$. Each access to the Oracle is treated as a separate IFO operation. The computational complexity is measured by the number of iterations each client performs on the IFO to reach the $\epsilon$-stationary point specified in Definition I. \\

%\textit{\textbf{Definition III}} (Incremental First-order Oracle (IFO)). An IFO is a subroutine that accepts an index $i \in [n]$ and a point $x \in \mathbb{R}^d$, and returns the pair $(f_i(x), \bigtriangledown f_i(x))$.

\par
Our objective is to find a critical point of F, where $\bigtriangledown F (\omega)=0$, i.e., to converge to an $\epsilon$-stationary point for smooth non-convex functions, i.e., $\mathbb E[\lVert \bigtriangledown f(x) \rVert^2] \le \epsilon$. Only we are trying to access stochastic gradient on arbitrary points. In this experiment, we consider stochastic gradient descent (SGD) as a standard algorithm. Using the following recursion, SGD generates a series of iterates $\omega_1, \omega_2, \cdots,\omega_N$.
\begin{equation}
    \omega_{i+1} = \omega_i - \eta_i g_i
\end{equation}
where $g_i=\bigtriangledown f(x_i, \omega_i), f(\cdot, x_1), \cdots, f(\cdot, x_N)$ are either the i.i.d. or non-i.i.d samples from a distribution $\mathcal{D}$. The learning rates $\eta_1,\eta_2, \cdots, \eta_N \in \mathbb{R}$ affect the performance if not tuned properly. The learning rate, i.e., step size plays a crucial role in the convergence of SGD. In order to address this sensitivity and render SGD resilient to parameter selection, ``adaptive'' SGD methods are frequently employed, in which the step-size is determined dynamically utilizing stochastic gradient data from both the current and previous samples \cite{Cutkosky2019, Khanduri2020distributed}. In this work, we present one such ``adaptive'' technique for distributed non-convex stochastic optimization that designs the step sizes based on the available stochastic gradient information. Proper selection of the learning rate, $\eta_i$ in SGD ensures that a randomly chosen recurrence $\omega_i$ ensures $\mathbb{E}[\lVert\bigtriangledown F(\omega_i)\rVert] \le O(1/N^{1/4})$. 

\subsection{Variance reduction}
The use of momentum-based variance reduction performs well in prior works to reduce the number of hyperparameters such as batch size. The momentum-based variance reduction is also used to reduce the variance of gradient in non-convex updates. In this work, the variance is represented as follows: \\

\begin{equation}
	\begin{aligned}
		m_{i} = (1-\beta)m_{i-1} + \beta{\bigtriangledown f(x_i, \omega_i)} \\ + (1-\beta)(\bigtriangledown f(x_i, \omega_i) - \bigtriangledown f(x_{i-1}, \omega_i)) \\
		\omega_{i+1} = \omega_i - \eta_i m_i
		\label{eq:vr}
	\end{aligned}
\end{equation} 
One additional term, $(1-\beta)(\bigtriangledown f(x_i, \omega_i) - \bigtriangledown f(x_{i-1}, \omega_i))$ is used to the update with adaptive learning rate. The variance reduction concept is similar to conventional reduction, where two different gradients are used in each step. 

\section{Proposed Algorithm}
To improve FL's convergence rate and communication overhead the following problems must be resolved: (i) the high variance of simple averaging used in the global server aggregation step when there are multiple local updates, which is made worse by client heterogeneity; (ii) the high variance linked to the noise of local client-level stochastic gradients; and the (iii) heterogeneity among the local functions. The key idea of the proposed FL algorithm is to apply momentum-based variance reduction for the global and local updates with the adaptive learning rate for clients. 
\begin{algorithm} [ht]
\caption{Algorithm for Client Update}
\label{alg:cup}
\textbf{Input}: $c$, $k$, $w$, Initial point:$\omega_1$
\begin{algorithmic}[1]
\STATE $\eta_{0} \leftarrow \frac{k}{w^{1/3}}$
\FORALL{$t=1,2,\cdots T$}
%\ForEach {$t=1,2,\cdots T$}
\FORALL {clients $i=1,2,\cdots N$ in parallel}
\FORALL {local epochs $j=1,2,\cdots E$}
  \IF {$j=1$}
      \STATE Set $m_i^{(t,j)}=\bigtriangledown{f_i(\omega_{i}^{(t,j)})}, \hat{m}_i^{(t-1,j)}=\bigtriangledown{f_i(\hat{\omega}_{i}^{(t-1,j)})}$
    \ELSE
      \STATE Each client $i$, randomly select a batch size, $\mathcal{B}_i^{(t,j)}$.
      \STATE Compute the stochastic gradient of the non-convex loss function, $f_i$ over $\mathcal{B}_i^{(t,j)}$ at $\omega_{i}^{(t,j)}$, $\hat{\omega}_{i}^{(t-1,j)}$, $\omega_{i}^{(t,j-1)}$, and $\hat{\omega}_{i}^{(t-1,j-1)}$, i.e., $\widetilde\nabla{f_i(\omega_{i}^{(t,j)};\mathcal{B}_i^{(t,j)})}$, $\widetilde\nabla{f_i(\omega_{i}^{(t-1,j)};\mathcal{B}_i^{(t,j)})}$, $\widetilde\nabla{f_i(\omega_{i}^{(t,j-1)};\mathcal{B}_i^{(t,j)})}$ and $\widetilde\nabla{f_i(\omega_{i}^{(t-1,j-1)};\mathcal{B}_i^{(t,j)})}$.
      \STATE $\eta_t \leftarrow \frac{k}{(w_t+\widetilde\nabla{f_i(\omega_{i}^{(t,j)})})^{1/3}}$
      \STATE Update $m_i^{(t,j)}=\widetilde\nabla{f_i(\omega_{i}^{(t,j)};\mathcal{B}_i^{(t,j)})} + m_i^{(t,j-1)}-\widetilde\nabla{f_i(\omega_{i}^{(t,j-1)};\mathcal{B}_i^{(t,j)})}$ and $m_i^{(t-1,j)}=\widetilde\nabla{f_i(\omega_{i}^{(t-1,j)};\mathcal{B}_i^{(t,j)})} + m_i^{(t-1,j-1)}-\widetilde\nabla{f_i(\omega_{i}^{(t-1,j-1)};\mathcal{B}_i^{(t,j)})}$ 
     \ENDIF
     \STATE Update $\omega_{i}^{(t,j+1)}=\omega_{i}^{(t,j)} - \eta_{t}m_i^{(t,j)}$ and $\hat{\omega}_{i}^{(t-1,j+1)}=\hat{\omega}_{i}^{(t-1,j)}- \eta_{t}\hat{\omega}_{i}^{(t-1,j)}$.
\ENDFOR
\ENDFOR
\ENDFOR
\STATE Send $(\omega^t - \omega_i^{t,E})$ and $((\omega^t - \omega_i^{t,E})-(\omega^{t-1} - \hat{\omega}_i^{t-1,E}))$ to the server.
\end{algorithmic}
\end{algorithm}
Hence, in the optimized federated non-convex heterogeneous settings, the total of $N$ number of clients are jointly trying to solve the following optimization problem:\\

\begin{equation}
 \min_{\omega \in \mathbb{R}}[F(\omega) := \sum_{i=1}^{N}S_{i}F_i(\omega)]
 \label{eq:p1}
\end{equation}
where $S_i = \frac{N_i}{N}$ is the relative sample size, $F_i(\omega) = \frac{1}{N_i}\sum_{x \in \mathcal{D}_i}f_i(x;\omega)$ is the $i^{th}$ client's local objective function. In our case, the learning model defined $f_i$ as a non-convex loss function and $x \in \mathcal{D}_i$ represents the data sample from the local data $\mathcal{D}_i$. After receiving the current global model $\omega^{(t,0)}$ at $t^{th}$ communication rounds, each client independently executes $\kappa_i$ iterations of the local solver to optimize its local objective. In our work, the local solver is the momentum-based variance reduction with an adaptive learning rate.
\\
The number of local updates $\kappa_i$ for each client $i$ can vary in our theoretical framework. If a client run $E$ epochs with batch size $\mathcal{B}$, then $\kappa_i = \floor{\frac{EN_i}{\mathcal{B}}}$, where $N_i$ is the sample data of the $i^{th}$ client. Recall from the simplest and most popular FL algorithm, FedAvg \cite{McMahan2017}, the update rule can be written as
follows:
\begin{equation}
    \omega^{(t+1,0)} - \omega^{(t,0)} =  \sum_{i=1}^N S_i\Delta_i^t = - \sum_{i=1}^N S_i. \eta \sum_{j=0}^{\kappa_i -1}g_i(\omega_i^{(t,j)})
 \end{equation}
where $\omega_i^{(t,j)}$ represents the model of the $i$-th client in the $t$-th communication rounds after the $j$-th local update, at round $t$, $\Delta_i^t$ is the local parameters changes of $i$-th client. In our work, we have modified the update rule where local updates are performed using momentum-based variance reduction with an adaptive learning rate, and the global update is performed using momentum-based variance reduction. The client and server update rules are demonstrated using the following algorithms.

\begin{algorithm} 
\caption{Algorithm for Server Update}\label{alg:sup}
\begin{algorithmic}[1]
\REQUIRE Initial parameter $\omega_1$, Number of communication rounds T, Number of participating clients $R \le N$, Epochs E
\ENSURE $\omega_0=\omega_1$
\FORALL {round $t=1,2,\cdots $}
\STATE Server broadcasts $\omega^t, \omega^{t-1}$ to a the set of $R$ number of clients, $\mathcal{S}_t$
\FORALL{client $i \in \mathcal{S}_t$}
\STATE Set $\omega_{i}^{(t,0)}=\omega_{t}$ and $\hat{\omega}_{i}^{(t-1,0)}=\omega_{t-1}$. Execute Algorithm $\ref{alg:cup}$.
\ENDFOR
\IF {$t=0$}
\STATE Set $\hat{m}^{(t)} = \frac{1}{R}\sum_{i \in \mathcal{S}^(t)}(\omega^{(t)}-\omega_i^{(t,E)})$ \COMMENT {$\hat{m}$ is the global momentum}
\ELSE
\STATE Set $\hat{m}^{(t)} = \frac{\beta^{(t)}}{R}\sum_{i \in \mathcal{S}^(t)}(\omega^{(t)}-\omega_i^{(t,E)})+(1-\beta^{(t)})\hat{m}^{(t-1)} + \frac{1-\beta^{(t)}}{R}\sum_{i \in \mathcal{S}^(t)}((\omega^{(t)}-\omega_i^{(t,E)}) -  (\omega^{(t-1)} - \hat{\omega}_i^{(t-1,E)}))$
 \ENDIF
 \STATE Update $\omega^{(t+1)} = \omega^{t}-\hat{m}^{(t)}$
\ENDFOR
\end{algorithmic}
\end{algorithm}

Cutosky and Orabona \cite{Cutkosky2019} first proposed the variance reduction method in non-convex settings and proposed an algorithm, namely STORM. In this paper, the idea of the STORM is considered as an update rule in a slightly different way. Equation \ref{eq:vr} represents the update rule in STORM. In equation \ref{eq:vr}, $\beta \in [0,1)$ is the momentum parameter per iteration and $x_i$ is the randomly selected sample in $i$-th iteration. The stochastic gradient at $\omega_{i+1}$ is calculated on $x_i$. In Algorithm \ref{alg:cup}, lines 6, 11, and 13 give the client updates using momentum-based variance reduction using adaptive learning rate after running $E$ number of epochs. It is pertinent to mention that the concept is similar to \ref{eq:vr} where $\beta=0$ and the full gradient is considered in the computation of the initial gradient. Similarly, in the Algorithm \ref{alg:sup}, global momentum is used for server aggregation. To overcome the pitfall of the used server aggregation scheme in conventional federated averaging such as FedAvg \cite{McMahan2017}, in this paper, we consider the gradient as $(\omega^{(t)}-\omega_i^{(t,E)})$.

%%%%%%%%%%%%%%%%%%%%%%%%%%%%%%%%%
\section{Empirical Analysis}
This section empirically studies the proposed method and compares it with different baseline methods to demonstrate its efficacy by applying it to solve federated learning problems
and image classification tasks. Experiments are implemented using PyTorch, and we run all experiments on CPU machines with 3.50 GHz Intel Xeon with NVIDIA GeForce RTX 2080 Ti GPU.
\subsection{Implementation Details}
\textbf{Dataset} Our image classification tasks will involve utilizing the MNIST dataset and CIFAR-10 dataset with 50 clients in the network. The MNIST dataset comprises 60,000 training images and 10,000 testing images, categorized into 10 classes. Each image is comprised of 28×28 arrays of grayscale pixels. The CIFAR-10 dataset encompasses 50,000 training images and 10,000 testing images, featuring 60,000 32×32 color images sorted into 10 categories. Each client will be equipped with an identical Convolutional Neural Network (CNN) model serving as the classifier, and the loss function employed will be cross entropy we use Dirichlet distribution to generate a non-IID dataset with $\alpha=0.5$\\
\textbf{Baselines} In our study we perform a comparative analysis of our approach with SOTA methods: FedAvg \cite{McMahan2017}, FedProx \cite{li2020federated}, FedNova\cite{Wang2021FedNova} and FedGLOMO \cite{Das2022faster}. FedAvg is one of the federated optimization approaches that is most frequently used among them. One could think of FedProx as a re-parametrization and generalization of FedAvg. FedNova is a quick error convergence normalized averaging method that removes objective inconsistency. FedGLOMO used the local and global update using momentum-based variance reduction with quantized messages to reduce communication complexity. The major difference between our proposed method and FedGLOMO is that no message quantization is used. We also use an adaptive learning rate strategy for local updates for early convergence. \\
\textbf{Network Architecture} In our experiment, the neural network is used with two different layers and the ReLU activation function. 
The size of the hidden layers is 600. We train the models by using the categorical cross-entropy loss with $\ell_2$-regularization. PyTorch's weight decay value is set to $1e-4$ to apply $\ell_2$-regularization.
\begin{table*}[ht]
	\centering
	\resizebox{\textwidth}{!}{
		\begin{tabular}{cccccccccc}
			\hline
			\multicolumn{4}{c}{\textbf{Mechanisms}} & \multicolumn{4}{c}{\textbf{Non-IID Data}} & Sample & Communication\\
			\cline{1-10}
		 Adaptive Learning & Momentum &  \multicolumn{2}{c}{Variance Reduction} & \multicolumn{2}{c}{MNIST} &\multicolumn{2}{c}{CIFAR-10} & &\\
			\hline
			 & & Local & Global & Accuracy(\%) & Loss & Accuracy(\%) & Loss\\
			\hline
		 X & X & X & X & $75.24 \pm 0.45$  & 0.67 & $72.45 \pm 0.34$ & 0.86 & $\mathcal{O}(\epsilon^ {-4} )$ & $\mathcal{O}(\epsilon^ {-4} )$\\
		\hline 
		 X & \checkmark & X & X & $90.19 \pm 0.40$  & 0.41 & $86.92 \pm 0.42$ & 0.73 & $\mathcal{O}(\epsilon^ {-4} )$ & $\mathcal{O}(\epsilon^ {-4} )$\\
		\hline 
		 \checkmark & X & X & X & $90.78 \pm 0.45$  & 0.20 & $87.05 \pm 0.45$ & 0.46 & $\mathcal{O}(\epsilon^ {-4} )$ & $\mathcal{O}(\epsilon^ {-4} )$\\
		\hline 
			\checkmark & \checkmark & X & X & $91.24 \pm 0.45$  & 0.0024 & $86.45 \pm 0.38$ & 0.0078 & $\mathcal{O}(\epsilon^ {-3} )$ & $\mathcal{O}(\epsilon^ {-3} )$\\
		\hline 
			 X & \checkmark & \checkmark & X & $89.24 \pm 0.45$  & 0.0026 & $85.32 \pm $ & 0.0032 & $\mathcal{O}(\epsilon^ {-4} )$ & $\mathcal{O}(\epsilon^ {-4} )$\\
		\hline 
		 X & \checkmark & \checkmark & \checkmark & $96.23 \pm 0.45$  & 0.0024 & $90.45 \pm 0.54$ & 0.0120 & $\mathcal{O}(\epsilon^ {-3} )$ & $\mathcal{O}(\epsilon^ {-1.5} )$\\
		\hline 
			 \checkmark & \checkmark & \checkmark & \checkmark & $98.04 \pm 0.40$  & 0.0017 & $92.12 \pm 0.35$ & 0.0020 & $\mathcal{O}(\epsilon^ {-3/2} )$ & $\mathcal{O}(\epsilon^ {-1} )$\\
		\hline 
		\end{tabular}
	}
	\caption{Ablation Studies with different components of the proposed method}
	\label{tab:abs}
\end{table*}

\begin{figure}[ht] 
	\centering 
	\noindent\begin{minipage}[t]{4cm} 
		\centering 
		\includegraphics[scale=0.35]{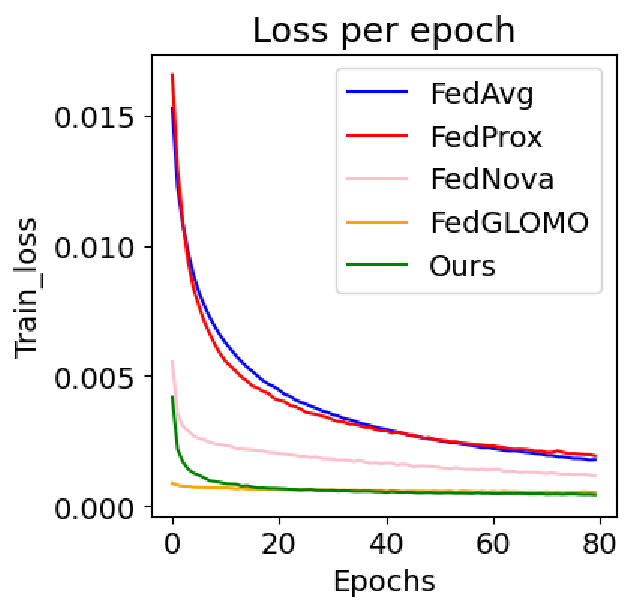} 
		\caption{Train Loss vs Epochs with CIFAR-10} 
  \label{fig:lossepoch}
	\end{minipage} 
	\vspace{3cm} 
	\noindent\begin{minipage}[t]{4cm}
		\centering 
		\includegraphics[scale=0.35]{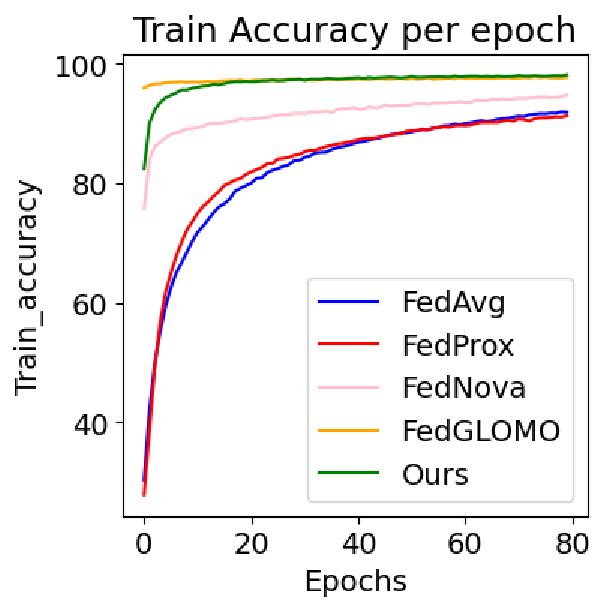} 
		\caption{Train Accuracy vs Epochs with CIFAR-10} 
  \label{fig:trainepoch}
	\end{minipage}
 \vskip -8\baselineskip  plus -1fil
 \end{figure}
 \begin{figure}[!h]
    \centering
    \begin{minipage}[t]{4cm} 
    \centering 
     \includegraphics[scale=0.35]{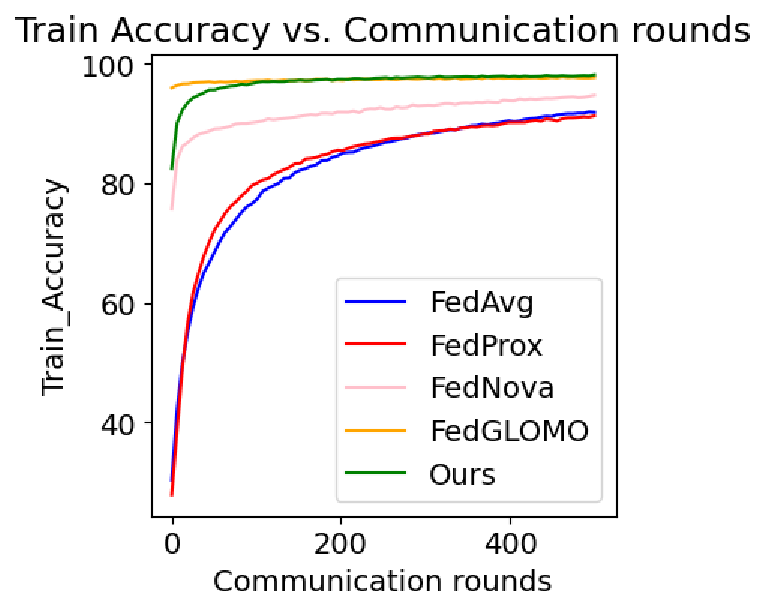}
    \caption{Train Accuracy vs Communications with CIFAR-10}
    \label{fig:traincomm}
    
\end{minipage} 
\vspace{3cm} 
	\begin{minipage}[t]{4cm} 
    \centering
    \includegraphics[scale=0.35]{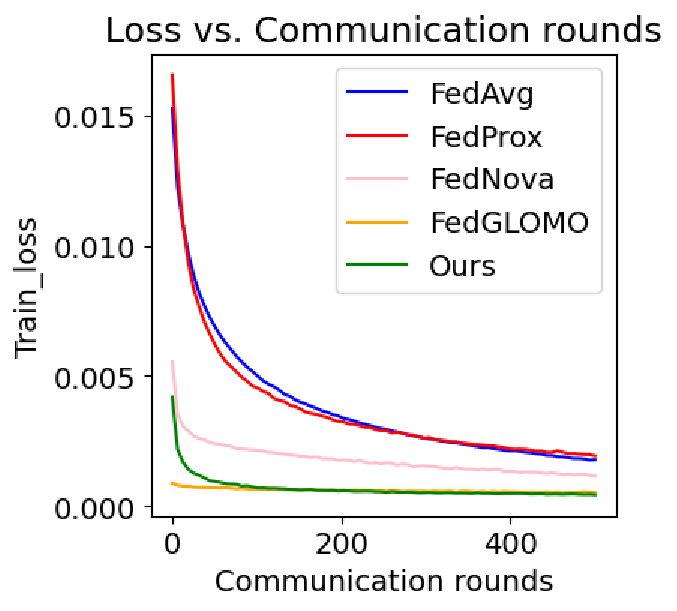}
    \caption{Train Loss vs Communications with CIFAR-10}
    \label{fig:losscomm}
    \end{minipage}
    \vskip -8\baselineskip  plus -1fil
\end{figure}
\begin{figure}[!h]
    \centering
    \begin{minipage}[t]{4cm} 
    \centering 
   \includegraphics[scale=0.35]{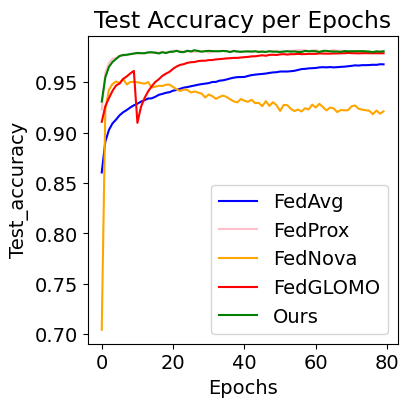}
    \caption{Test Accuracy with CIFAR-10}
    \label{fig:testepoch}
\end{minipage}
\vspace{3cm} 
	\begin{minipage}[t]{4cm} 
    \centering
    \includegraphics[scale=0.35]{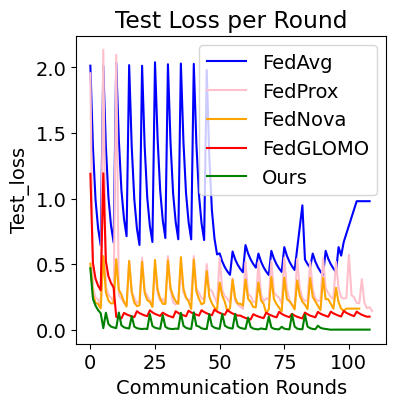}
    \caption{Test Loss with CIFAR-10}
    \label{fig:testcomm}
    \end{minipage}
    \vskip -8\baselineskip  plus -1fil
\end{figure}
\subsection{Validation}
\begin{figure}[ht] 
	\centering 
	\noindent\begin{minipage}[t]{4cm} 
		\centering 
		\includegraphics[scale=0.35]{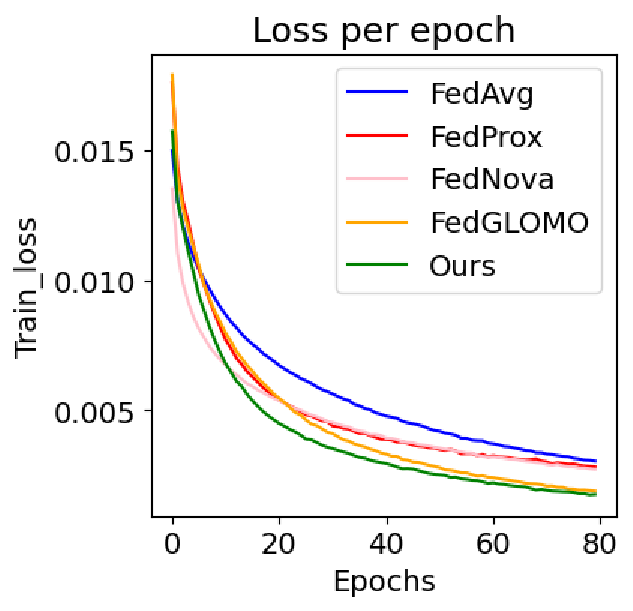} 
		\caption{Train Loss vs Epochs with MNIST} 
  \label{fig:lossepochmnist}
	\end{minipage} 
	\vspace{3cm} 
	\noindent\begin{minipage}[t]{4cm}
		\centering 
		\includegraphics[scale=0.35]{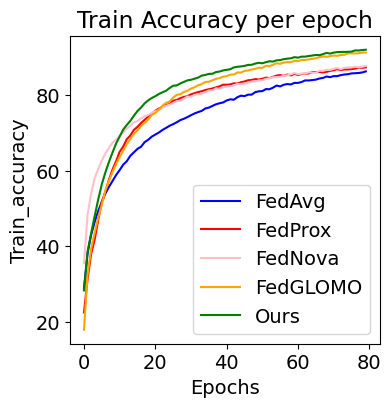} 
		\caption{Train Accuracy vs Epochs with MNIST} 
  \label{fig:trainepochmnist}
	\end{minipage}
 \vskip -8\baselineskip  plus -1fil
 \end{figure}

\begin{figure}[!h]
    \centering
    \begin{minipage}[t]{4cm} 
    \centering 
    \includegraphics[scale=0.35]{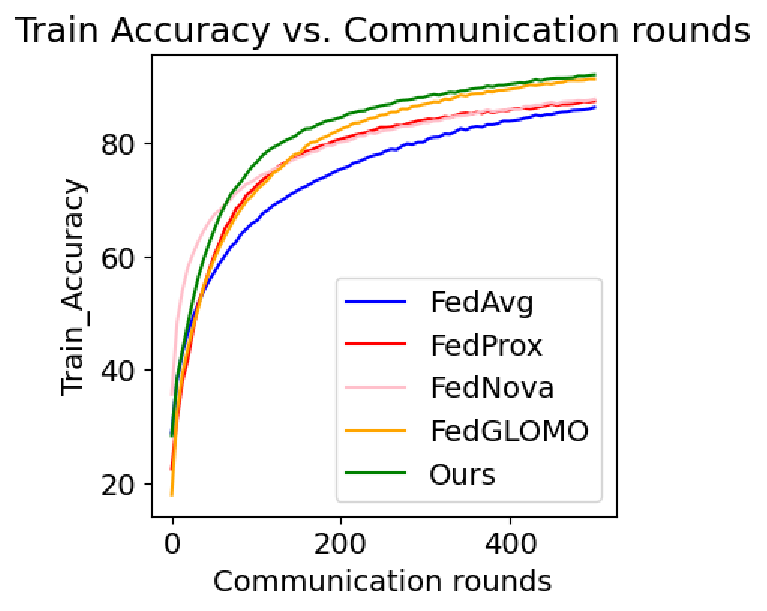}
    \caption{Train Accuracy vs Communications with MNIST}
    \label{fig:traincommmnist}
    
\end{minipage} 
\vspace{3cm} 
	\begin{minipage}[t]{4cm} 
    \centering
    \includegraphics[scale=0.35]{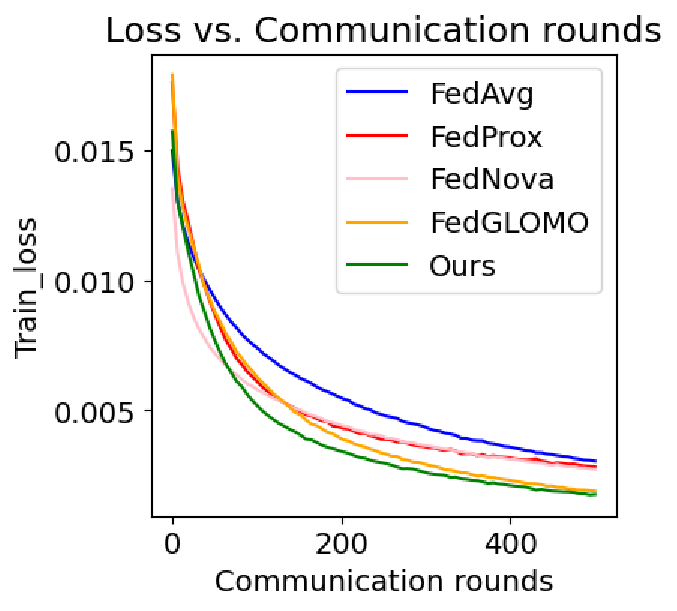}
    \caption{Train Loss vs Communications with MNIST}
    \label{fig:losscommmnist}
    \end{minipage}
    \vskip -8\baselineskip  plus -1fil
\end{figure}

\begin{figure}[!h]
    \centering
    \begin{minipage}[t]{4cm} 
    \centering 
    \includegraphics[scale=0.35]{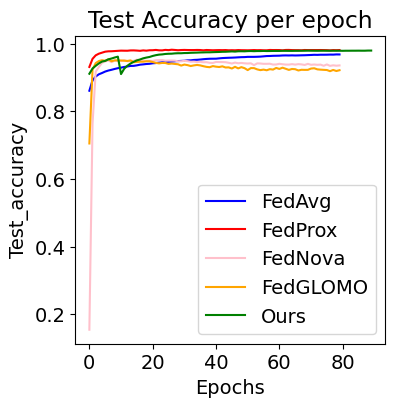}
    \caption{Test Accuracy with MNIST}
    \label{fig:testepochmnist}
\end{minipage}
\vspace{3cm} 
	\begin{minipage}[t]{4cm} 
    \centering
    \includegraphics[scale=0.35]{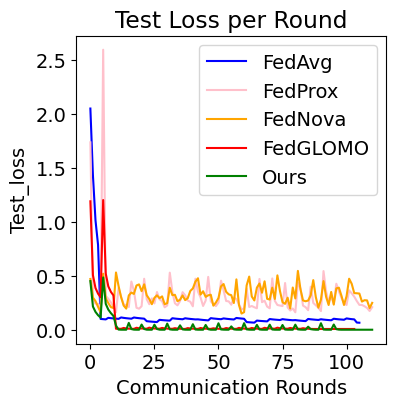}
    \caption{Test Loss with MNIST}
    \label{fig:testcommmnist}
    \end{minipage}
    \vskip -8\baselineskip  plus -1fil
\end{figure}

Figures \ref{fig:lossepoch} and \ref{fig:trainepoch} demonstrate the cross-entropy-based training loss and accuracy with the CIFAR-10 dataset concerning the number of epochs on the federated clients. Similarly, Figures \ref{fig:losscomm} and \ref{fig:traincomm} show the training loss and accuracy with the CIFAR-10 dataset concerning the number of communication rounds. Moreover, Figures \ref{fig:testepoch} and \ref{fig:testcomm} demonstrate the test accuracy and loss in each communication round. The test loss and accuracy clearly depict the early convergence of our proposed method. 
Similarly, Figures \ref{fig:lossepochmnist} and \ref{fig:trainepochmnist} demonstrate the cross-entropy-based training loss and accuracy with the MNIST dataset concerning the number of epochs on the federated clients. Similarly, Figures \ref{fig:losscommmnist} and \ref{fig:traincommmnist} show the training loss and accuracy with the MNIST dataset concerning the number of communication rounds. Figures \ref{fig:testepochmnist} and \ref{fig:testcommmnist} represent the test accuracy and loss using the MNIST dataset and also prove the early convergence.
Using both the datasets, the experimental results depict that, while our proposed FL algorithm is marginally better than AdaGLOMO on test accuracy, on both the training loss and accuracy our proposed algorithm appears to be somewhat faster in terms of the number of communication rounds. Moreover, the testing accuracy is more stable than the SOTA methods as the variations of the accuracy are less in our proposed method. The test accuracy and loss results depict the early convergence of the proposed method. 
\subsection{Ablation Studies}
In the proposed FL algorithm we use local updates with adaptive learning and global updates. Both the local and global updates use momentum-based variance reduction. Table \ref{tab:abs} shows the performance with different components such as SGD with adaptive learning, momentum-based variance reduction for local updates, and global updates. For this experiment, we have used 80 epochs with 400 rounds and a batch size of 50. Out of 100 clients, a random number of clients participated in each round. It should be noted that the size of the batch used in our experiment is small. This confirms the effectiveness of the proposed method in overcoming the need for large batch sizes for convergence.
 
%%%%%%%%%%%%%%%%%%%%%%%%%%%%%%%%%%%%%%%%%%%%%
\section{Conclusion}
In this study, a communication-efficient approach to federated learning is proposed, leveraging both momentum-based variance reduction and adaptive learning rates. Both local and global updates integrate momentum-based variance reduction and adaptive learning rates for the local models, thereby reducing communication cycles for early convergence. The proposed method exhibits robust performance in non-convex heterogeneous settings with non-IID data, surpassing several state-of-the-art methods. The principal aim of this paper is to minimize loss, a demonstration accomplished through thorough empirical and theoretical analyses.
\par
The proposed method is experimented on a cross-device FL framework. While performing the theoretical analysis, several assumptions are made to prove the convergence. Here, the belief that all the clients participate to show the reduced complexity is a major limitation of this work. In the future, we will come up with cross-silo FL with a similar strategy. 

\clearpage
\bibliography{aaai25}

\newpage
\appendix

\section{Appendix / Supplemental material}
In this section, we present the convergence analysis and proof of the convergence of the proposed federated learning algorithm. 

%%%%%%%%%%%%%%%%%%%%%%%%%%%%%%%%%%%%%%%%%%%%%%%%%%
\subsection{Convergence Analysis}
The convergence is proved using the following theorems.
\begin{theorem}
Based on the Assumptions \ref{asu1}, \ref{asu2}, \ref{asu3}, and \ref{asu4} and for initial batch size  $\mathcal{B} =bE$, we set $\eta_t=\frac{k}{(w_t+\sigma^2t)^{1/3}}$, 
$k=\frac{(bN)^{2/3}\sigma^{2/3}}{L}$. Also set $c = \frac{(8L)^2}{bN} + \frac{\sigma^2}{24LEk^3} = L^2(\frac{64}{bN}+\frac{1}{24(bN)^2E})$ and $w = max\{(4LkE)^3-\sigma^2t, 2\sigma^2, \left( \dfrac{ck}{16LE}\right)^3\}$. Then \\
we can set the local updates, E and batch size, b as follows:\\
\begin{equation}
    E=\mathcal{O}\left((T/N^2)^{v/3} \right), \\ 
    b=\mathcal{O}\left((T/N^2)^{1/2 - v/2} \right)
\label{eq:Eb}    
\end{equation}
where $v \in [0,1]$.

\end{theorem}
After applying the variance reduction with the adaptive learning rate for noise removal, according to Theorem 1, we have 
\begin{equation}
\begin{split}
          \mathbb{E}\lVert \bigtriangledown {f(\omega)} \rVert^2 = \mathcal{O}\left(\frac{f(\omega_1) - f^*}{N^{2v/3}T^{1-v/3}} \right) + \mathcal{\tilde{O}}\left(\frac{\sigma^2}{N^{2v/3}T^{1-v/3}} \right) + \\ \mathcal{\tilde{O}}\left(\frac{\bigtriangledown^2}{N^{2v/3}T^{1-v/3}} \right)
          \end{split}
\end{equation}
For any $v \in [0, 1]$, the sample complexity of the proposed method is $\mathcal{\tilde{O}}\left( \epsilon^{-3/2} \right)$. Hence, each client involved in a communication cycle required at most $\mathcal{\tilde{O}}\left( N^{-1}\epsilon^{-3/2} \right)$ gradient computations. Moreover, the communication complexity is $\mathcal{\tilde{O}}\left( \epsilon^{-1} \right)$.
\par
Before going to the proof of the above convergence analysis, We explain some of the tradeoffs as follows: \\
The tradeoff between sample size and communication complexity: From the above theorem, the sample and communication complexities are represented as $\mathcal{\tilde{O}}\left( \epsilon^{-3/2} \right)$ and $\mathcal{\tilde{O}}\left( \epsilon^{-1} \right)$ when E and b are selected efficiently. In the FL literature, we find the logarithmic factor complexity using either irrespective of the sample or batch Lipschitz smooth assumption. Hence, our FL framework outperforms in terms of sample and communication complexities. Based on our assumptions the $\mathcal{\tilde{O}}\left( \epsilon^{-1} \right)$ is the optimal complexity in comparison with SOTA methods.\\
The tradeoff between the batch sizes and local updates: The number of local updates E and the batch sizes b are balanced using the parameters $v \in [0,1]$. The equations in \ref{eq:Eb} demonstrate the relation between E and b with the interval [0,1]. As $v$ grows from o to 1 the batch size b decreases and the local updates E increases. At $v=1$, the b is $\mathcal{O}(1)$, however, $E=\mathcal{O}(\frac{T^{1/3}}{N^{2/3}})$. Conversely, at $v=0$, $b=\mathcal{O}(\frac{T^{1/2}}{N})$, however, E is $\mathcal{O}(1)$. These concepts generalize the SGD and minibatch SGD with local and global updates using momentum-based variance reduction.

\par

Before going to the details of the proof, it is necessary to discuss some lemmas in detail.
\subsection{Preliminary Lemmas}
\begin{lemma}\label{lemma1}
	We can define the error term as $\epsilon_t = m_t - \frac{1}{N}\sum_{n=1}^N \bigtriangledown f^{(k)}(\omega_t^{n})$, then the iterations according to Algorithm \ref{alg:cup} \\
	\begin{dmath*}
		\mathbb{E} \left[ \left\langle (1 - \beta_t)\epsilon_{t-1}, \frac{1}{N}\sum_{n=1}^{N}\frac{1}{b}\sum_{x_t^{n} \in \mathcal{B}_t^{(n)}} \left[ \left( \bigtriangledown f^{(n)} \left( \omega_t^{n};x_t^{(n)} \right) \nonumber\\ - f^{(n)} (\omega_t^{n} ) \right)  - (1-\beta_t) \left( \bigtriangledown f^{(n)} (\omega_{t-1}^{n}; x_t^{(n)}) - \bigtriangledown f^{(n)} ( \omega_{t-1}^{n}) \right) \right]  \right\rangle \right] = 0
		\label{eq:12}
	\end{dmath*}
	
\end{lemma}
\begin{proof}
	Let for some value of N, the gradient error term is $\epsilon_{t-1}$. For all $n \in N$, the only randomness in the left half of the Lemma statement w.r.t $x_t^{n}$. This suggests that we've \\
	\begin{dmath*}
		\mathbb{E} \left[ \left\langle (1 - \beta_t)\epsilon_{t-1}, \frac{1}{N}\sum_{n=1}^{N}\frac{1}{b}\sum_{x_t^{n} \in \mathcal{B}_t^{(n)}} \left[ \left( \bigtriangledown f^{(n)} ( \omega_t^{n};x_t^{(n)}) \\ - f^{(n)} ( \omega_t^{n} ) \right) - (1-\beta_t) \left( \bigtriangledown f^{(n)} (\omega_{t-1}^{n}; x_t^{(n)}) - \bigtriangledown f^{(n)} ( \omega_{t-1}^{n}) \right) \right]  \right\rangle \right]
		= \mathbb{E} \left[ \left\langle (1 - \beta_t)\epsilon_{t-1}, \frac{1}{N}\sum_{n=1}^{N}\frac{1}{b}\sum_{x_t^{n} \in \mathcal{B}_t^{(n)}} \left[ \left( \bigtriangledown f^{(n)} ( \omega_t^{n};x_t^{(n)}) - \\ f^{(n)} ( \omega_t^{n} ) \right)  - (1-\beta_t) \left( \bigtriangledown f^{(n)} (\omega_{t-1}^{n}; x_t^{(n)}) - \bigtriangledown f^{(n)} ( \omega_{t-1}^{n}) \right) \right] | \mathcal{F}_t \right\rangle \right] 
		\label{eq:13}
	\end{dmath*}
	where $\mathcal{F}_t=\sigma(\omega_1^{n}, \omega_2^{n},\cdots, \omega_t^{n})$ for all $n \in [N]$.
	If $x_t^{n}$ is chosen randomly with uniform distribution at each $k \in K$ and 
	$\mathcal{E}[\bigtriangledown f^{n}(\omega^(n); x_t^{k})] = \bigtriangledown f^(n)(\omega_t^{(n)})$, then we have\\
	\begin{dmath*}
		\mathbb{E} \left[ \frac{1}{b} \sum_{x_t^{t} \in \mathcal{B}_t^{(n)}} \left[ \left( \bigtriangledown f^{(n)} (\omega_t^{n}; x_t^{n}) - \bigtriangledown f^{(n)} (\omega_t^{(n)}) \right)  - (1-\beta_t) \left( \bigtriangledown f^{(n)} (\omega_{t-1}^{n}; x_{t-1}^{n}) - \bigtriangledown f^{(n)} (\omega_t^{(n)})\right)   \right] | \mathcal{F}_t \right]=0
	\end{dmath*}
	for all $n \in N$.\\
	Hence, proved.
\end{proof}

\begin{theorem}
	Choosing the parameters as 
	
	\begin{itemize}
		\item $k=\frac{(bN)^{2/3}\sigma^{2/3}}{L}$
		\item $c = \frac{(8L)^2}{bN} + \frac{\sigma^2}{24LEk^3} = L^2(\frac{64}{bN}+\frac{1}{24(bN)^2E})$ 
		\item $w = max\{(4LkE)^3-\sigma^2t, 2\sigma^2, \left( \dfrac{ck}{16LE}\right)^3\}$
	\end{itemize}
	and for any $v \in [0,1]$ at each client and the total number of local updates  $E=\mathcal{O}\left((T/N^2)^{v/3} \right)$, \\ 
	batch size $b=\mathcal{O}\left((T/N^2)^{1/2 - v/2} \right)$ and the initial batch size, $\mathcal{B} = bE$, the proposed FL algorithms satisfies the following:\\
	\begin{enumerate}[label=(\roman*)]
		\item $\mathbb{E}\lVert \bigtriangledown {f(\omega)} \rVert^2 = \mathcal{O}\left(\frac{f(\omega_1) - f^*}{N^{2v/3}T^{1-v/3}} \right) + \mathcal{\tilde{O}}\left(\frac{\sigma^2}{N^{2v/3}T^{1-v/3}} \right) + \mathcal{\tilde{O}}\left(\frac{\bigtriangledown^2}{N^{2v/3}T^{1-v/3}} \right)$.
		\item Sample complexity: To reach the $\epsilon$-stationary point the proposed FL algorithm requires at most $\mathcal{O}(\epsilon^{-3/2})$ gradient computations.Hence, each client requires at most $\mathcal{O}(N ^ {-1}\epsilon^{-3/2})$ gradient computations.
		\item Computation Complexity: To reach the $\epsilon$-stationary point the proposed FL algorithm requires at most $\mathcal{O}(\epsilon^{-1})$ communication rounds.
	\end{enumerate}
\end{theorem}
\begin{proof}
	\begin{itemize}
		\item Proof of Statement (i): Put the values of $\mathcal{B}$, E, and b in the given expression and replace $\mathcal{B}=bE$, we get \\
		$\mathcal{E}\lvert \bigtriangledown f(\omega) \rvert ^ 2 \le [ \frac{32LE}{T}  + \frac{2L}{(bN)^{2/3}T^{2/3}}](f(\omega_1)-f^*) + [\frac{8E}{T}+ \frac{1}{2(bN)^{2/3}T^{2/3}}]\sigma^2 + [\frac{256^2E}{T}+ \frac{64^2}{(bN)^{2/3}T^{2/3}}]\sigma^2log(T+1)  + [\frac{256^2E}{T}+ \frac{64^2}{(bN)^{2/3}T^{2/3}}]\bigtriangledown^2 \frac{E-1}{E}log(T+1)$
		Considering the fact that the total number of local updates  $E=\mathcal{O}\left((T/N^2)^{v/3} \right)$, \\ 
		batch size $b=\mathcal{O}\left((T/N^2)^{1/2 - v/2} \right)$ we will get the expression mentioned in (i).
		\item Sample Complexity: From the above expression, the total number of required iterations to achieve $\epsilon$-stationary point is as follows: \\
		$ \mathcal{O}(\frac{1}{N^{2v/3}T^{1-v/3}}) = \epsilon \implies T=\mathcal{O}(\frac{1}{N^{2v/(3-v)}\epsilon^{3/(3-v)}})$.
		Each client computes 2b stochastic gradients in each iteration. Hence, each client computes $2bT$ number of iterations in total. Using $b=\mathcal{O}\left((T/N^2)^{1/2 - v/2} \right)$, at each client, the number of total gradient computations is as follows:\\
		$bT = \mathcal{O}(\frac{T^{3/2-v/2}}{N^{1-v}}) = \mathcal{O}(\frac{1}{N\epsilon^{3/2}})$\\
		Therefore, the sample complexity is $\mathcal{O}(\epsilon{-3/2})$
		\item Communication Complexity: To achieve $\epsilon$-stationary point, the total number of communication rounds are $T/B$, with $B=\mathcal{O}\left((T/N^2)^{v/3} \right)$ and the $T=\mathcal{O}(\frac{1}{N^{2v/(3-v)}\epsilon^{3/(3-v)}})$, we get the communication complexity as follows:\\
		$T/B = \mathcal{O}(T^{1-v/3}N^{2v/3}) = \mathcal{O}(\frac{1}{\epsilon})$
	\end{itemize}
	Hence, the theorem.
\end{proof}

\end{document}